\documentclass{article}

\usepackage{PRIMEarxiv}
\usepackage[utf8]{inputenc} 
\usepackage[T1]{fontenc}    
\usepackage{hyperref}       
\usepackage{url}            
\usepackage{booktabs}       
\usepackage{amsfonts}       
\usepackage{nicefrac}       
\usepackage{microtype}      
\usepackage{lipsum}
\usepackage{fancyhdr}       
\usepackage{graphicx}       
\graphicspath{{media/}}     
\usepackage{svg}
\usepackage{authblk}
\usepackage[paper=portrait,pagesize]{typearea}
\usepackage{multirow} 
\usepackage{makecell}

\usepackage{geometry}
\usepackage{multirow}
\usepackage{booktabs}
\usepackage{makecell}
\usepackage{pdflscape}

\usepackage{svg}
\usepackage{enumitem}
\usepackage{caption}
\usepackage{bm}

\usepackage{float}
\usepackage{amsmath}

\usepackage{xcolor,pifont}
\newcommand*\colourcheck[1]{%
  \expandafter\newcommand\csname #1check\endcsname{\textcolor{#1}{\ding{52}}}%
}
\colourcheck{blue}
\colourcheck{green}
\colourcheck{red}
\colourcheck{orange}

\usepackage{array}
\usepackage{ragged2e}

\usepackage{titlesec}
\titlespacing{\section}{0pt}{\parskip}{-\parskip}
\titlespacing{\subsection}{0pt}{\parskip}{-\parskip}
\titlespacing{\subsubsection}{0pt}{\parskip}{-\parskip}

\usepackage{tcolorbox}
\usepackage{enumitem}

\usepackage{amsmath}
\usepackage{algorithm}
\usepackage{algpseudocode}
\usepackage{amsfonts}
\usepackage{amsmath}
\usepackage{graphicx}
\usepackage{hyperref}

\usepackage{graphicx}
\DeclareGraphicsExtensions{.png}
\pdfpagewidth=\paperwidth
\pdfpageheight=\paperheight

\usepackage{graphicx}
\usepackage{float}
\usepackage{caption}

\usepackage{graphicx,float,caption}

\usepackage{apacite}
\usepackage{natbib}

\usepackage{lipsum}
\usepackage{graphicx}
\graphicspath{{media/}}     
\usepackage{float}
\usepackage{authblk}

\usepackage{authblk}
\title{From Staff Messages to Actionable Insights: A Multi-Stage LLM Classification Framework for Healthcare Analytics}

\author[ ]{\textbf{Hajar Sakai}, \textbf{Yi-En Tseng}, \textbf{Mohammadsadegh Mikaeili}, \textbf{Joshua Bosire}, and \textbf{Franziska Jovin}}
\affil[ ]{Cooper University Health Care}
\affil[ ]{Camden, NJ, USA}

\begin{document}
\maketitle

\begin{abstract}
Hospital call centers serve as the primary contact point for patients within a hospital system. They also generate substantial volumes of staff messages as navigators process patient requests and communicate with the hospital offices following the established protocol restrictions and guidelines. This continuously accumulated large amount of text data can be mined and processed to retrieve insights; however, traditional supervised learning approaches require annotated data, extensive training, and model tuning. Large Language Models (LLMs) offer a paradigm shift toward more computationally efficient methodologies for healthcare analytics. This paper presents a multi-stage LLM-based framework that identifies staff message topics and classifies messages by their reasons in a multi-class fashion. In the process, multiple LLM types, including reasoning, general-purpose, and lightweight models, were evaluated. The best-performing model was o3, achieving 78.4\% weighted F1-score and 79.2\% accuracy, followed closely by gpt-5 (75.3\% Weighted F1-score and 76.2\% accuracy). The proposed methodology incorporates data security measures and HIPAA compliance requirements essential for healthcare environments. The processed LLM outputs are integrated into a visualization decision support tool that transforms the staff messages into actionable insights accessible to healthcare professionals. This approach enables more efficient utilization of the collected staff messaging data, identifies navigator training opportunities, and supports improved patient experience and care quality.
\end{abstract}


\section{Introduction}
The growing complexity of healthcare delivery, driven by fragmented care systems, rising data volumes, and increasing patient expectations, demands intelligent and integrated solutions that enhance patient care and streamline clinical and non-clinical workflows (HIMSS, 2020). Serving as the front line of hospital services, a typical healthcare facility's Patient Contact Center (PCC) is the first point of contact in any patient's journey. Therefore, mining the data continuously generated within a PCC has great potential for leading to data-driven decision-making. When a patient contacts a hospital, a navigator or agent picks up and handles the call from within a PCC. During the call, depending on the patient's questions and needs, an encounter message might be created. This message is used for documenting the call and sending a message to different pools, usually representing different entities in the hospital (e.g., ambulatory offices).
\\[0.2cm] 
Large volumes of staff messages are generated daily, and reading them all to get business insights is unrealistic. Encounter messages are largely unstructured text, extracting actionable insights poses challenges related to classification accuracy, annotation costs, and semantic ambiguity (Hossain et al., 2023). On the one hand, given the resource constraints in healthcare settings and the manual effort required to label textual data, traditional text mining and supervised learning-based machine learning models come with challenges in scalability and adaptability (Mascio et al., 2020; Ghassemi et al., 2018; Brown III et al., 2021). On the other hand, discovering the information that remains buried in unstructured text would result in not only decision support for message senders (i.e. navigator or agent) and recipients, but also process and quality improvement in patient care by providing an automated solution leveraging visualization tools and LLMs. This study aims to address both real operational needs and technical innovation. 
\\[0.2cm] 
This paper involves the development of a data-driven decision support tool aimed at improving processes that involve patients, PCC, and internal hospital entities. The tool emphasizes data mining, visualization, and usability while accommodating users with diverse professional backgrounds and varying levels of data literacy. The end goal is to provide users with the multidimensional information needed to support their different decision-making scenarios (e.g., staffing, training) in an aggregated and summarized format, instead of requiring them to navigate through the raw data. In addition to visualizing staff messaging structured data through a Power BI dashboard, the encounter messages are analyzed through prompt engineering-based topic modeling and hierarchical text multi-class classification to identify the different reasons behind sending them by leveraging an LLM. The proposed solution involves two main Microsoft tools, Power BI and Azure, offering an accessible decision support tool while using cloud-based services and ensuring HIPAA compliance. Figure 1 overviews the research problem emanating from the healthcare professionals' need to ameliorate the efficiency of the presented process and improve the communication workflow.

\begin{figure}[H] \centering \includegraphics[scale=0.65]{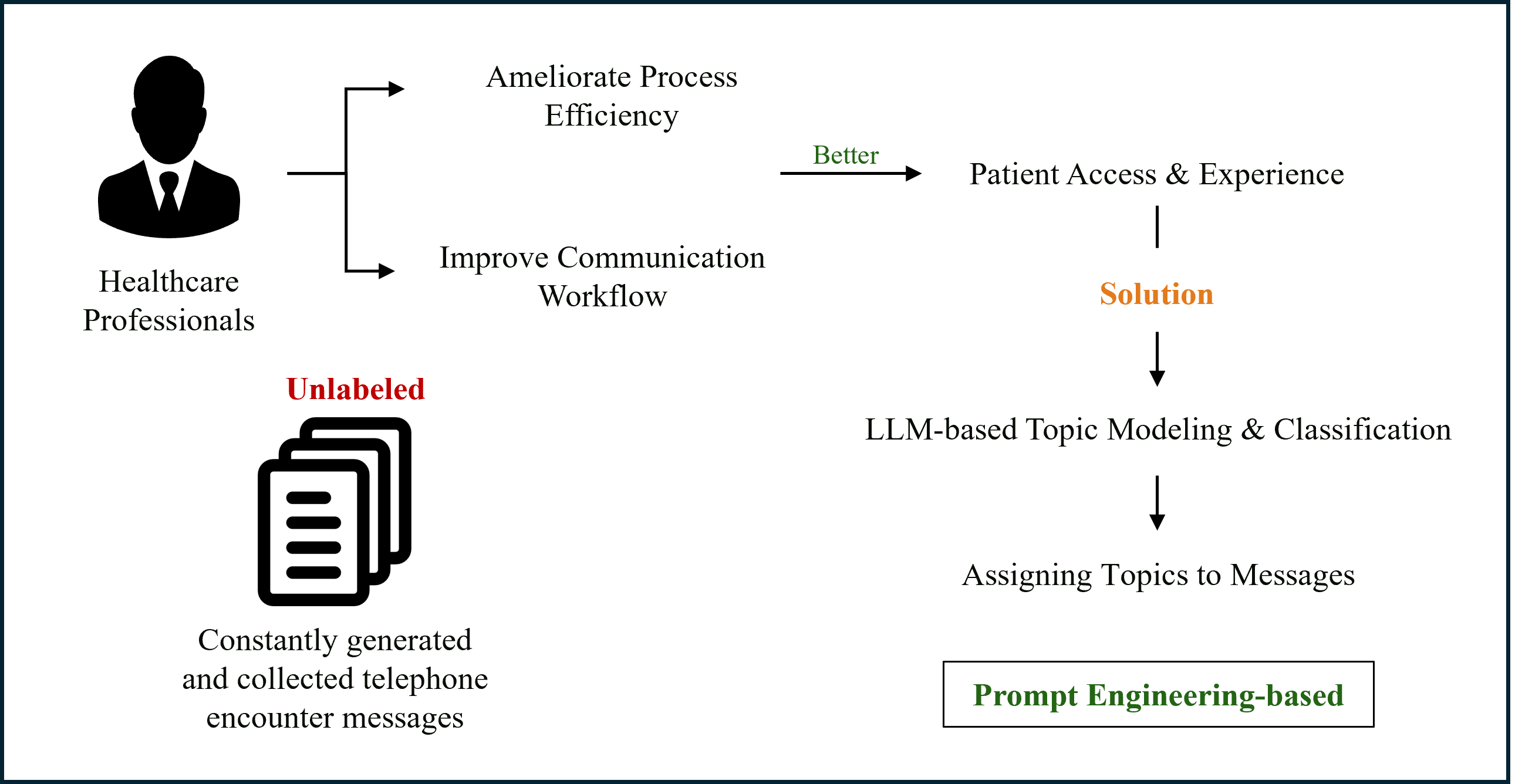} \caption*{\textbf{Figure 1.} Research Problem} \end{figure} 

As detailed in Figure 2, this paper contributes to the paradigm shift that healthcare is experiencing by combining LLM-based text analysis with real-time visualization, transforming complex text analysis into actionable insights for healthcare professionals and leading to improved patient care and operational efficiency. 

\begin{figure}[H] \centering \includegraphics[scale=0.65]{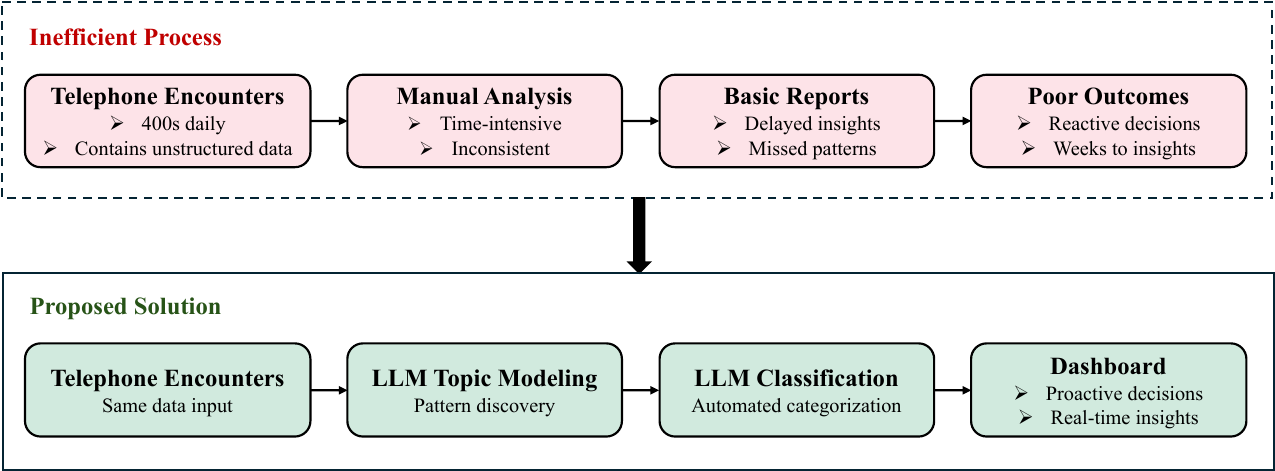} \caption*{\textbf{Figure 2.} Contribution} \end{figure} 

This research is structured as follows: The "Related Work" section reviews literature on LLM-based text mining and data visualization. The "Data and Methodology" section describes the dataset used and introduces the proposed multi-stage methodology encompassing topic modeling, text classification, and decision support tool development. The results and final outputs are discussed in the "Results and Discussion" section. Finally, the "Conclusion and Future Directions" are presented.

\section{Related Work}
Encounter messages generated during hospital operations present both a challenge and an opportunity for healthcare analytics. The introduction of visualization decision support tools like dashboards has enabled the conversion of regularly produced raw data into insights that contribute to the decision-making process, and the advent of LLMs has been transformative for such applications. This section examines the current state of research involving LLMs for healthcare text classification, as well as the use of visualization tools for developing data-driven processes in healthcare settings.
\\[0.2cm]
Visual analytics and dashboards have gained prominence as critical tools in health systems engineering, enabling rapid decision-making and effective communication in data-intensive environments. In particular, decision support tools like dashboards play an essential role in synthesizing complex and fragmented healthcare data into accessible, actionable formats for diverse stakeholders, reducing cognitive load using visual aids (Khan et al., 2017) and surpassing the limitations of static reports. They support not only data visualization but also workflow efficiency, behavioral change, and user engagement (Zhuang et al., 2022). As it is the case in this paper, staff messages sent internally are a critical part of the patient access workflow, yet often neglected and the literature reveals that decision support tools like dashboard integration with LLMs to tackle unstructured text problem suffer from scarcity, leaving room for contribution. 
\\[0.2cm]
The healthcare industry, unlike other domains, is reputed for being slower in adopting AI (He et al., 2019). Despite the transformative potential of AI in general, and generative AI more specifically in healthcare, their widespread integration is lagging due to multiple constraints, including regulation and HIPAA compliance, as well as healthcare professionals' trust and resistance.
\\[0.2cm]
Traditionally, messages conveyed and shared within hospitals are either categorized manually or by adopting a sequential flow consisting of first preprocessing the collected text, followed by converting text into vectors or embeddings before feeding them into machine learning models (e.g., Support Vector Machine, Decision Tree), all in a supervised learning fashion. For instance, Heisey-Grove et al. (2021) categorized secure electronic messages between patients and clinicians by reading and assigning classes while involving coding agreement analysis among researchers. Kalabikhina et al. (2024) classified patient reviews about medical services written in Russian by sentiment (i.e., positive or negative) and by target (i.e., physician, clinic, or mixed). The authors proposed a hybrid methodology consisting of using the Gated Recurrent Unit (GRU) for classification and Named Entity Recognition (NER), referred to as the linguistic component. Sulieman et al. (2017) focused on classifying patient portal messages into four categories (i.e., informational, medical, social, and logistical). Their best-performing approach consisted of Word2Vec for embeddings, a Convolutional Neural Network (CNN) architecture, all using a one-versus-all binary classification approach. Klimis et al. (2021) used two types of text data: outgoing messages and replies, with the former classified into five intents and the latter into seven categories. A 10-fold cross-validation for overfitting handling and grid search for hyperparameter tuning were leveraged with DistilBERT for text embeddings and L2-regularized Logistic Regression for the classification.
\\[0.2cm]
The manual and supervised approaches are characterized by multiple time-consuming and resource-intensive challenges, including continuous manual data labeling and model training and validation. The emergence of LLMs has offered great assistance to most NLP tasks, and text classification is no exception. General-purpose language models have also been proven to be efficient in accurately classifying contextual data in healthcare settings.  developed a fusion framework that combines three different types of pretrained Bidirectional Encoder Representations from Transformers (BERT) models (i.e., RoBERTa, BioBERT, and BERTweet) using a Convolutional Neural Network (CNN) layer to categorize the patient portal messages into the corresponding primary concerns. As a result, these LLMs were used to address the growing challenge of efficiently triaging the increasing volume of patient portal messages in healthcare. Chen et al. (2025) introduced an optimization algorithm that combines Kalman filtering with Adaptive Moment Estimation to improve the fine-tuning efficiency and accuracy of LLMs for medical sentiment analysis. The classification used a transformer-based classification approach in which a fine-tuned GPT4ALL model with BERT embeddings performs medical sentiment analysis on patient health feedback text for sentiment classification into positive, negative, or neutral. Sakai et al. (2024) used GPT-4 Turbo with prompt engineering to perform multi-label text classification of patient comments, testing 0-shot, 1-shot, 3-shot, and 5-shot learning scenarios where the model classified each comment into one or more of 10 predefined healthcare topics. Cho et al. (2024) compared traditional supervised learning models (i.e., Long Short Term Memory (LSTM) and BERT) against GPT-4's in-context learning capabilities using 0-shot and few-shot approaches to classify cancer patient call transcripts into five main categories. GPT-4 few-shot achieved the highest accuracy without requiring model retraining in categorizing complex and ambiguous patient inquiries from hospital call center transcripts. Batugo et al. (2024) leveraged GPT-4 with prompt engineering to perform binary classification of medication-related messages from a hypertension management program, comparing 0-shot, 1-shot, and few-shot learning approaches, where GPT-4 classified each message as belonging or not belonging to specific medication intent categories. GPT-4 was also used for prompt refinement.
\\[0.2cm]
Staff messages are critical yet often overlooked touchpoints in the patient journey that shape satisfaction and trust in the healthcare system (Arias et al., 2023). Many patients' complaints stem from poor communication, lack of empathy, and insufficient information—often first experienced during telephone interactions (Jangland et al. (2009)). These negative encounters cause anxiety and erode confidence in care. Therefore, leveraging LLMs and existing decision support tools could enable real-time reaction and proactive decision-making.

\section{Data and Methodology}

\subsection{Data}
This research integrates data from institutional databases and reference files to evaluate staff messaging and PCC activity within patient access operations. Staff messages data is extracted from the hospital's system and includes both encounter-level and encounter message-level tables. To identify the responsible staff and departments, employee information obtained from a human resources database is leveraged, restricted to individuals affiliated with the PCC and active. Supplemental mappings from two internal reference files are used to associate encounter records with ambulatory office locations. Call center activity data is sourced from the enterprise call management system and includes detailed records of inbound calls handled by PCC staff. These records are linked to agent identifiers and team affiliations to support analysis of call volume and staffing patterns. In this paper, experiments are run on a dataset of size 2,000 from which a subset of 500 messages was manually labeled for experiment evaluation. 

\subsection{Proposed Methodology}
In this paper, an end-to-end approach is proposed that transforms daily staff messages into actionable insights using LLMs through prompt engineering and a dashboard tool. A comprehensive evaluation of state-of-the-art LLMs is conducted to enable the selection of the most suitable model for this specific use case.

\begin{figure}[H]
    \centering
    \includegraphics[scale=0.75]{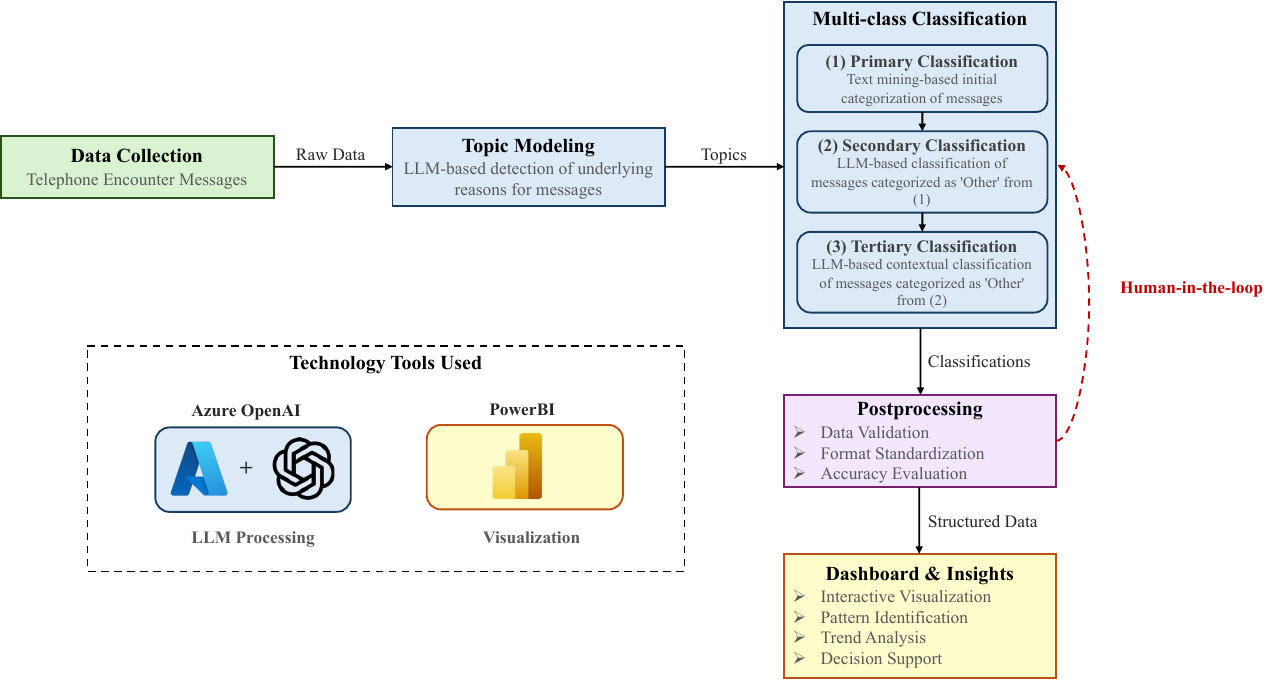}
    \caption*{\textbf{Figure 3.} Proposed Methodology}
\end{figure}

\subsubsection{Topic Modeling}

The staff messages collected represent the raw data; however, this data is characterized by not being associated with any topics or labels. Since the main goal of the classification is to shed light on the reasons behind sending these messages to thereafter improve the process's efficiency by uncovering areas of training improvement, an LLM (i.e., GPT-4o) was first used to identify the underlying reasons. This was conducted within Azure Foundry AI chat playground by adding a large set of encounter messages of size 1,000 as external knowledge to the LLM following a Retrieval Augmented Generation (RAG) system. By chatting with the LLM through prompt crafting, brainstorming internally, and implementing feedback, the hierarchy of topics summarized in Figure 4 was deduced. The resulting topics are well defined and representative of the data used, in addition to being relevant and providing information to healthcare professionals.

\begin{figure}[H]
    \centering
    \includegraphics[scale=0.45]{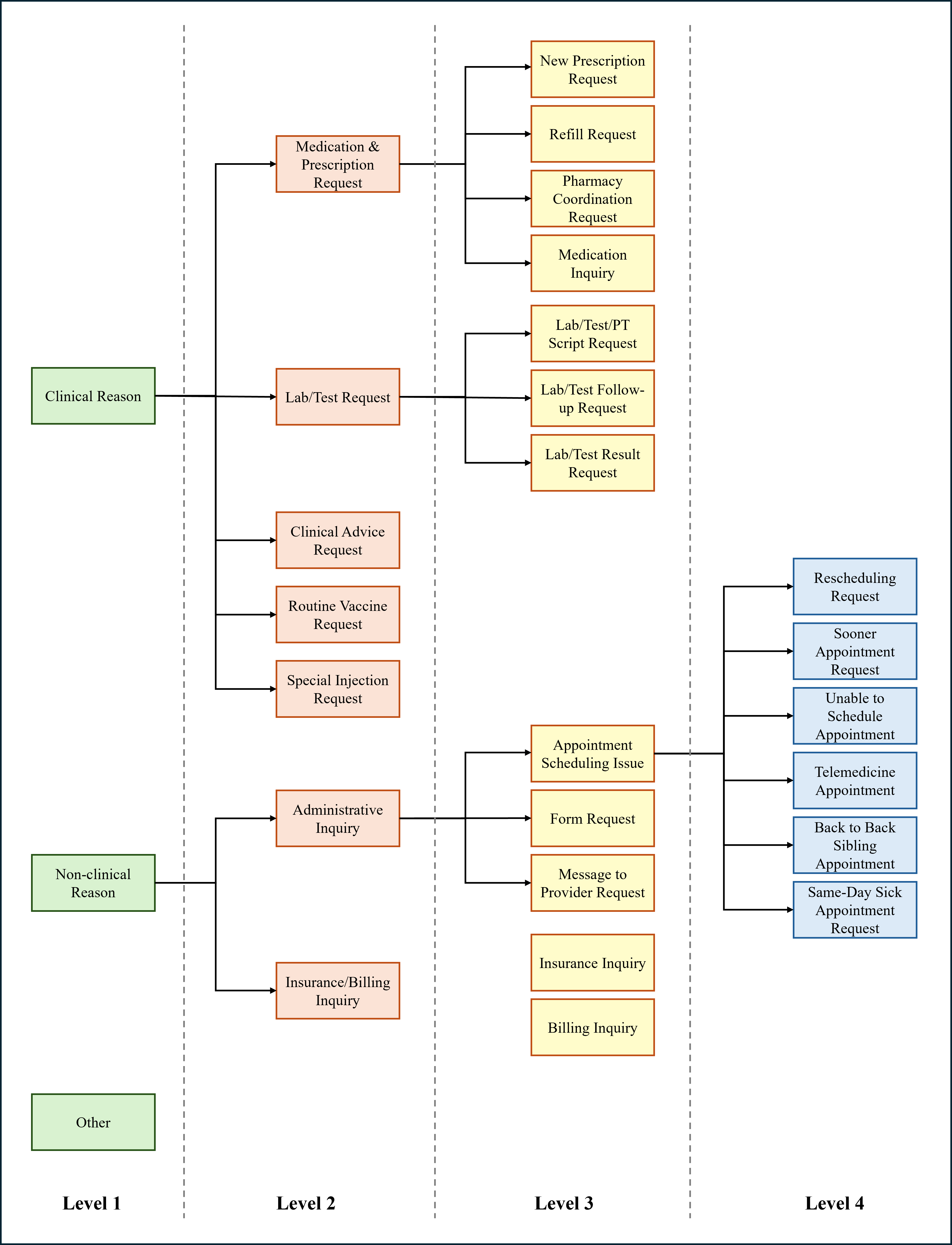}
    \caption*{\textbf{Figure 4.} Topics' Hierarchy}
\end{figure}

\subsubsection{Three-stage Classification}
Figure 4 presents the hierarchy of topics that reveals the logic behind the identification of topics. However, for visualization purposes, the most granular topic would be the most relevant. This transforms the classification from a hierarichal multi-label one to a multi-class one. In order to balance both computation efficiency and cost effectiveness, a three stage multi-class classification is proposed. 17 different LLMs available through Azure OpenAI service were deployed and evaluated: 

\begin{table}[h]
\centering
\caption*{\textbf{Table 1.} LLMs Considered Overview}
\begin{tabular}{c c c}
\hline
\textbf{LLM Type} & \textbf{LLM} & \textbf{Model Name} \\
\hline
\multirow{9}{*}{OpenAI GPT} & \multirow{3}{*}{Large/Flagship Models} & gpt-5 \\
\cline{3-3}
 & & gpt-4.1 \\
\cline{3-3}
 & & o3 \\
\cline{2-3}
 & \multirow{3}{*}{Mini Models} & gpt-5-mini \\
\cline{3-3}
 & & gpt-4.1-mini \\
\cline{3-3}
 & & o4-mini \\
\cline{2-3}
 & \multirow{2}{*}{Nano Models} & gpt-5-nano \\
\cline{3-3}
 & & gpt-4.1-nano \\
\cline{2-3}
 & Open Source & gpt-oss-120b \\
\hline
\multirow{2}{*}{Grok} & Large Model & grok-3 \\
\cline{2-3}
 & Mini Model & grok-3-mini \\
\hline
Meta Llama & Large Instruct Model & Llama-3.3-70B-Instruct \\
\hline
\multirow{2}{*}{DeepSeek} & Reasoning Model & DeepSeek-R1 \\
\cline{2-3}
 & Large Model & DeepSeek-V3 \\
\hline
\multirow{2}{*}{Microsoft Phi} & Standard Model & Phi-4 \\
\cline{2-3}
 & Mini Instruct Model & Phi-4-mini-instruct \\
\hline
Microsoft \& OpenAI & Routing System & model-router \\
\hline
\end{tabular}
\end{table}

Let $\mathcal{M} = \{m_1, m_2, \ldots, m_n\}$ be the set of staff messages to be classified.
\begin{itemize}
    \item \textbf{Stage 1: Primary Classification} \\
The first classification function $f_1$ performs initial text mining-based categorization based on keywords and key-phrases identified manually:
\begin{equation} f_1: \mathcal{M} \rightarrow \mathcal{C}_1 \cup \{\text{Other}_1\} \end{equation}
where: \begin{itemize} \item $\mathcal{C}_1 = \{c_1^1, c_2^1, \ldots, c_k^1\}$ represents the primary topics' set \item $\text{Other}_1$ denotes messages not classified into any topic at this stage \end{itemize} 
\vspace{0.5em}
This partitions the encounters' message set into: 
\begin{itemize} \item $\mathcal{M}_1 = \{m \in \mathcal{M} \mid f_1(m) \in \mathcal{C}_1\} \quad \text{(i.e., successfully classified messages)}$ \item $\mathcal{O}_1 = \{m \in \mathcal{M} \mid f_1(m) = \text{Other}_1\} \quad \text{(i.e., unclassified messages)}$ \end{itemize}
\vspace{0.5em}

    \item \textbf{Stage 2: Secondary Classification} \\
The secondary classification function $f_2$ applies LLM-based classification with prompt engineering $P_2$ to the remaining messages:
\begin{equation} f_2: \mathcal{O}_1 \rightarrow \mathcal{C}_2 \cup \{\text{Other}_2\} \end{equation}
where: \begin{itemize} \item $\mathcal{C}_2 = \{c_1^2, c_2^2, \ldots, c_j^2\}$ represents the secondary topics' set \item $P_2$ denotes the crafted prompt for LLM multi-class classification \item $\text{Other}_2$ denotes messages still unclassified even after the second classification stage \end{itemize}
\vspace{0.5em}
This further partitions $\mathcal{O}_1$ into: 
\begin{itemize} 
\item $\mathcal{M}_2 = \{m \in \mathcal{O}_1 \mid f_2(m; P_2) \in \mathcal{C}_2\} \quad \text{(i.e., classified messages at stage 2)}$ \item $\mathcal{O}_2 = \{m \in \mathcal{O}_1 \mid f_2(m; P_2) = \text{Other}_2\} \quad \text{(i.e., unclassified messages)}$ \end{itemize}
\vspace{0.5em}

    \item \textbf{Stage 3: Tertiary Classification} \\
The tertiary classification function $f_3$ applies contextual LLM-based classification with enhanced prompt engineering $P_3$, where the entire encounter message thread is fed to the LLM instead of just the first message:
\begin{equation} f_3: \mathcal{O}_2 \rightarrow \mathcal{C}_3 \cup \{\text{Other}_3\} \end{equation}
where: \begin{itemize} \item $\mathcal{C}_3 = \{c_1^3, c_2^3, \ldots, c_\ell^3\}$ represents the tertiary topics' set \item $P_3$ denotes the contextual prompt incorporating the entire message's thread \item $\text{Other}_3$ denotes the final set of unclassified messages \end{itemize}

\end{itemize}

As a result, the final multi-class classification output is obtained as follows:
\begin{equation} 
F(m) = \begin{cases} 
f_1(m) & \text{if } f_1(m) \in \mathcal{C}_1 \\
f_2(m; P_2) & \text{if } f_1(m) = \text{Other}_1 \land f_2(m; P_2) \in \mathcal{C}_2 \\
f_3(m; P_3) & \text{if } f_1(m) = \text{Other}_1 \land f_2(m; P_2) = \text{Other}_2 \land f_3(m; P_3) \in \mathcal{C}_3 \\
\text{Other}_3 & \text{otherwise} 
\end{cases} 
\end{equation}

\section{Results and Discussion}

\subsection{LLM Classification Performance Evaluation}

The comprehensive evaluation of 17 different LLMs for staff messaging multi-class classification reveals significant performance variations depending on the model architectures, sizes, and tasks trained on. The results reported in Figure 5 represent the evaluation of the proposed three-stage approach depending of the LLM considered. The experiments considers a dataset of size 2,000 at stage 1, 1,335 at stage 2 and a variable sample set size at stage 3 depending on the LLM, while the evaluation focuses on a subset of size 500 manually labeled. The experiments conducted demonstrate that model selection critically impacts classification accuracy, with performance ranging from 46.4\% to 78.4\% in terms of weighted F1-scores and from 40.8\% to 79.2\% in terms of accuracy as detailed in Figure 5.

\begin{figure}[H]
    \centering
    \includegraphics[scale=0.3]{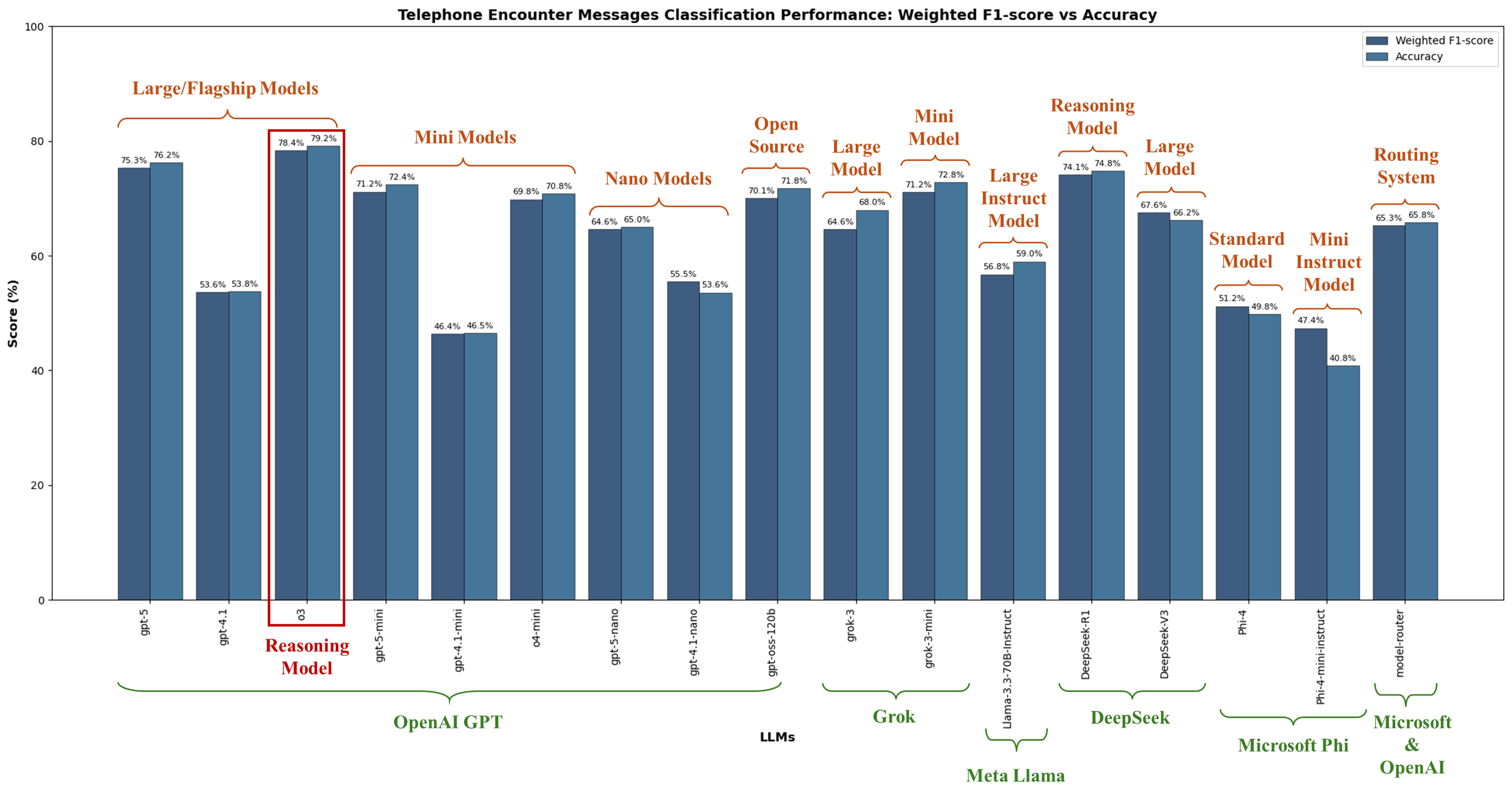}
    \caption*{\textbf{Figure 5.} Metrics Summary}
\end{figure}

On one hand, the top 3 performing LLMs are dominated by OpenAI's flagship offerings as well as DeepSeek-R1, with the o3 reasoning model achieving the highest weighted F1-score of 78.4\% and close accuracy of 79.2\%. gpt-5 follows with a weighted F1-score of 75.3\% and accuracy of 76.2\% and then DeepSeek-R1 with a weighted F1-score of 74.1\% and accuracy of 74.8\%, establishing these three LLMs as the current state-of-the-art for this classification task. These results suggest that the reasoning capabilities and advanced architecture of large flagship models provide substantial advantages for understanding the nuanced language patterns in staff messages.
\\[0.2cm]
On the other hand, when examining LLM size variants within the same families, a significant gap emerges, except for the case of grok-3 and grok-3-mini where the smaller model surprisingly outperforms the larger one. For the cases of GPT models and Phi models, the model scale remains a critical factor in achieving high performance when classifying staff messaging data, which brings up the potential trade-off of between computational efficiency and accuracy. 
\\[0.2cm]
Open-source models like grok-3 and gpt-oss-120b achieve respectable performance, especially the latter, which demonstrates that closed-source models don't have an exclusive advantage. However, Meta's Llama-3.3-70B-Instruct achieves a less satisfactory performance, while DeepSeek-V3 shows similar mid-tier performance. Therefore, for practical applications, hospitals have viable alternatives if expensive flagship models cannot be resorted to.
\\[0.2cm]
The lower-performing models, including Microsoft's Phi variants and Microsoft's model router, leveraging at the time the experiment was run gpt-4.1, gpt-4.1-mini, gpt-4.1-nano, and o4-mini where the optimal LLM is selected dynamically depending on the prompt and task to ensure accuracy and output quality while minimizing the cost, achieve more modest results. The substantial nearly 40\% gap between the best and worst performing LLMs, in terms of accuracy, highlights the significant differences in capability given the task complexity.
\\[0.2cm]
These results pinpoint the importance of carefully matching LLM selection to specific use case requirements. In healthcare applications like staff messaging classification, where accuracy has significant implications, the superior performance of flagship models (i.e., o3, gpt-5, and DeepSeek-R1) may justify their potential higher computational costs. However, for less critical applications, the less performing models might provide an optimal balance of performance and efficiency.
\\[0.2cm]
This inference time analysis of stage 2, summarized in Figure 6, reveals a significant performance gap between the fastest and slowest models, with processing times ranging from around 10 minutes for efficient models like Phi-4-mini-instruct to over 90 minutes for DeepSeek-R1. The results highlight a critical trade-off for hospitals deploying LLMs in production environments. While the slower, more capable models may deliver superior classification accuracy, as shown in the previous results, their extended processing times could make them impractical for real-time applications or high-volume processing scenarios. Fortunately, for the case of the staff messaging multi-class classification, the best-performing LLM, o3, is also characterized by significantly fast computation time.

\begin{figure}[H]
    \centering
    \includegraphics[scale=0.55]{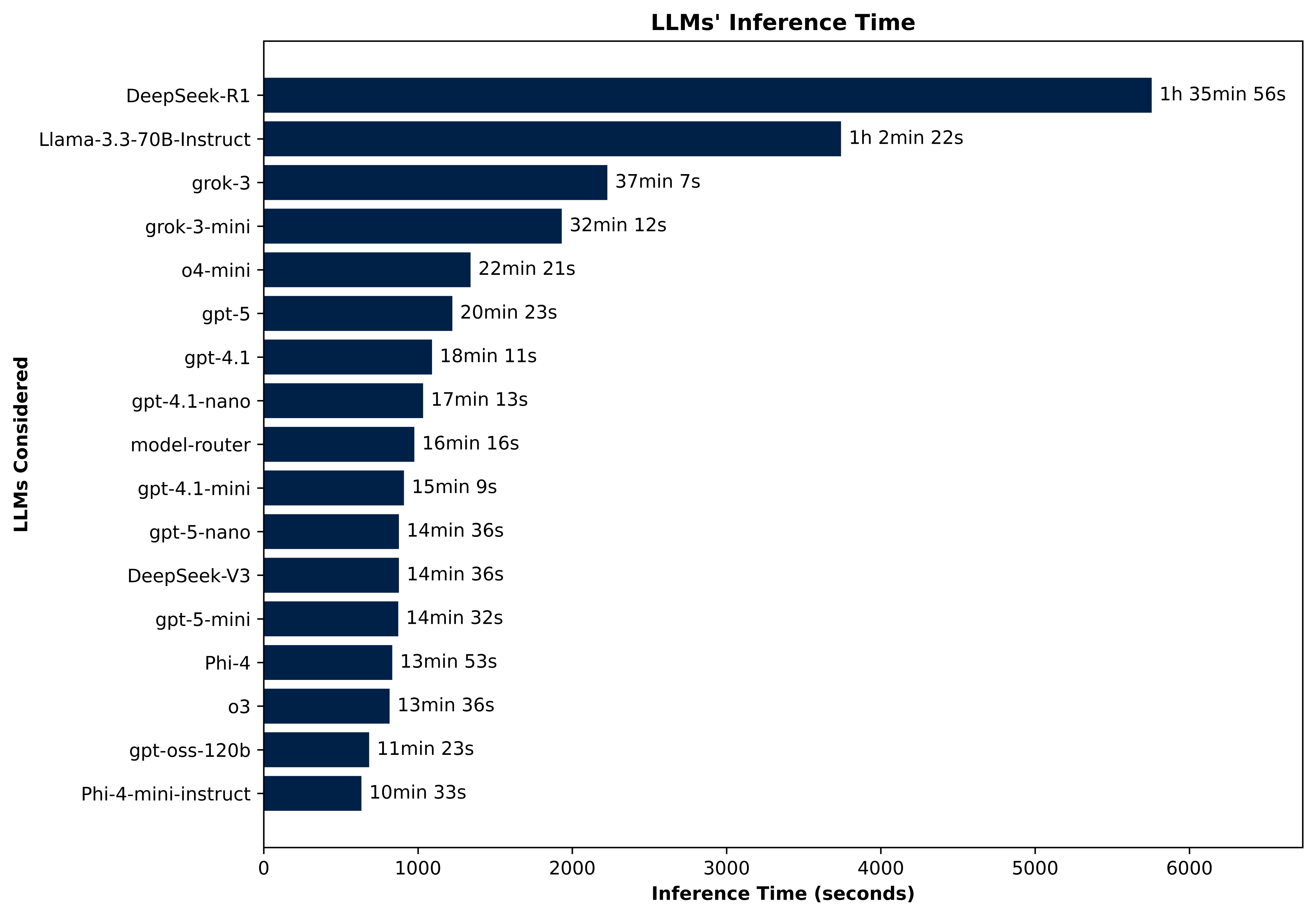}
    \caption*{\textbf{Figure 6.} Inference Time Summary}
\end{figure}

It is important to keep in mind that when deploying LLMs for multi-class classification of staff messaging, careful consideration of the risks of bias and hallucination could significantly impact decision-making accuracy. LLMs are prone to exhibiting systematic biases, potentially leading to misclassification that can have disproportionate effects. Additionally, these models can hallucinate classifications by confidently assigning categories that don't accurately reflect the message content, especially when dealing with ambiguous text data, which is often the case here. This underscores the importance of maintaining human oversight to ensure that the proposed automated multi-class classification system enhances rather than compromises fair and accurate assessment of staff communications displayed in a dashboard, providing actionable insights.

\subsection{Decision Support Tool and User Insights}

The PCC decision support tool provides an integrated visualization platform that transforms staff messaging data into actionable insights for healthcare operations and diverse stakeholders. The high-level overview highlights volumes, conversion rates, and routing distributions across pools, teams, and offices. These metrics allow healthcare professionals to quickly monitor contact center throughput, identify where patient requests are concentrated, and how workload is distributed across the organization. Temporal trend charts further reveal seasonal patterns or sustained shifts in encounter activity, offering managers the ability to anticipate surges and allocate resources more effectively.
\\[0.2cm]
The unique contribution of this system lies in its incorporation of LLM-classified staff messaging topics, which unlock a deeper understanding of patient needs. Within the decision support tool, a second page is dedicated for the staff messages. This page segments unstructured patient communications into clinical and non-clinical reasons (i.e., Figure 4), surfacing content-specific patterns that may otherwise remain hidden. Clinical subtopics (e.g., prescription refills, vaccine requests, lab results, or clinical advice) highlight demand on providers and nursing staff, pointing to areas where clinical protocols or targeted staffing could reduce bottlenecks. Equally important, non-clinical subtopics (e.g., same-day sick appointment requests, rescheduling, insurance inquiries, and billing questions) reveal operational challenges that directly impact patient access and satisfaction. The team-level and office-level topic distributions also provide comparative insights: which teams are managing higher proportions of clinical-related requests, or which offices are receiving specialized patient requests. Figures 7-10 provide snippets from the developed decision support tool. It is worth noting that for the sake of a simplified visualization, only the level 1 and the most granular subtopic are displayed in the decision support tool.

\begin{figure}[htbp]
    \centering
    \includegraphics[scale=0.405]{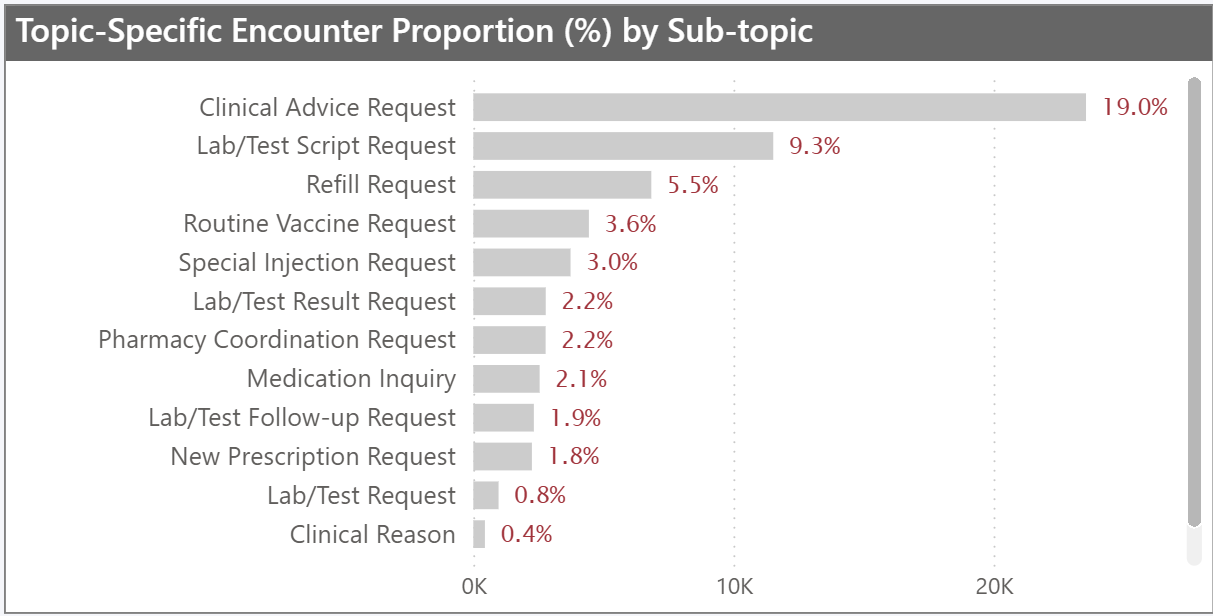}
    \caption*{\textbf{Figure 7.} Clinical Topic Distribution}
\end{figure}

\begin{figure}[H]
    \centering
    \includegraphics[scale=0.6]{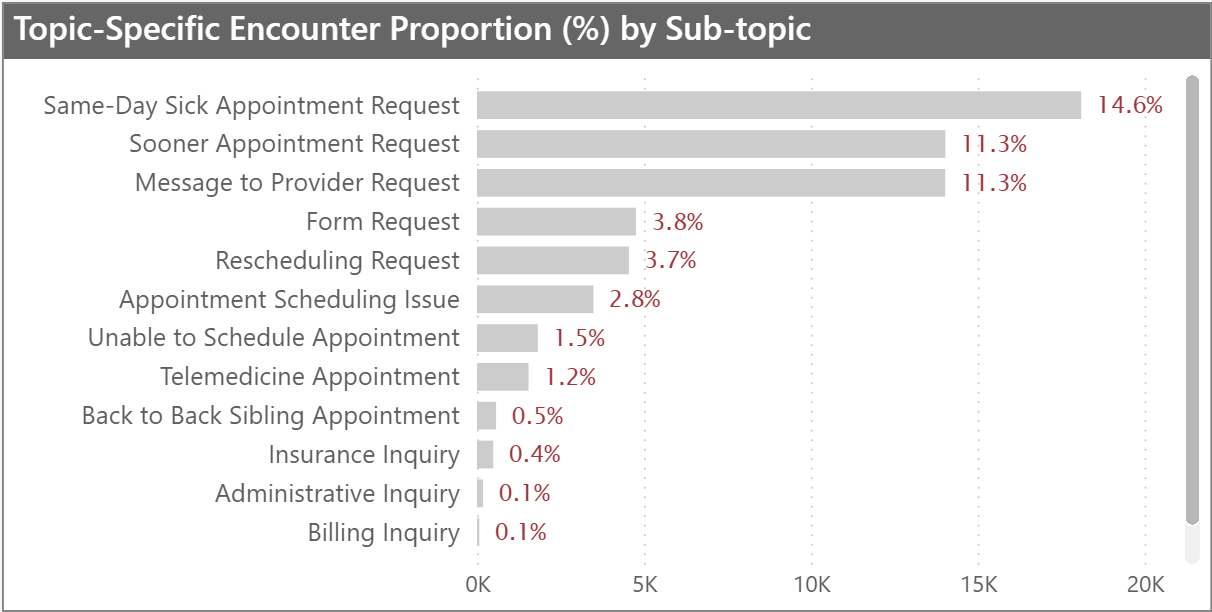}
    \caption*{\textbf{Figure 8.} Non-clinical Topic Distribution}
\end{figure}

\begin{figure}[H]
    \centering
    \includegraphics[scale=0.405]{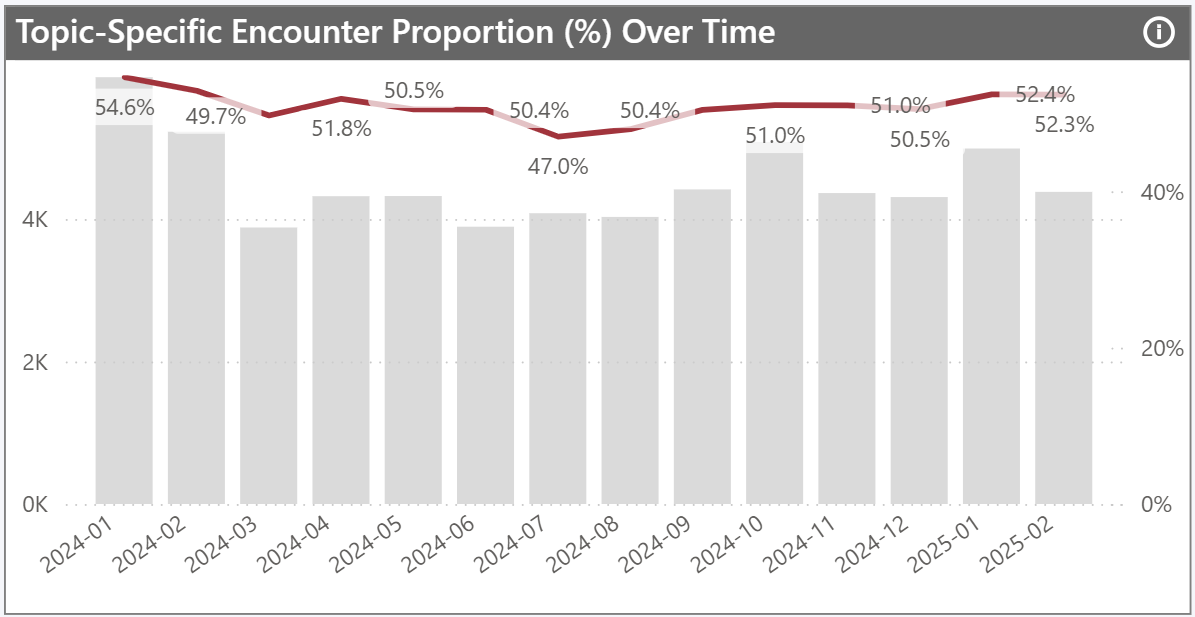}
    \caption*{\textbf{Figure 9.} Clinical Reason Topic Distribution Over Time}
\end{figure}

\begin{figure}[H]
    \centering
    \includegraphics[scale=0.55]{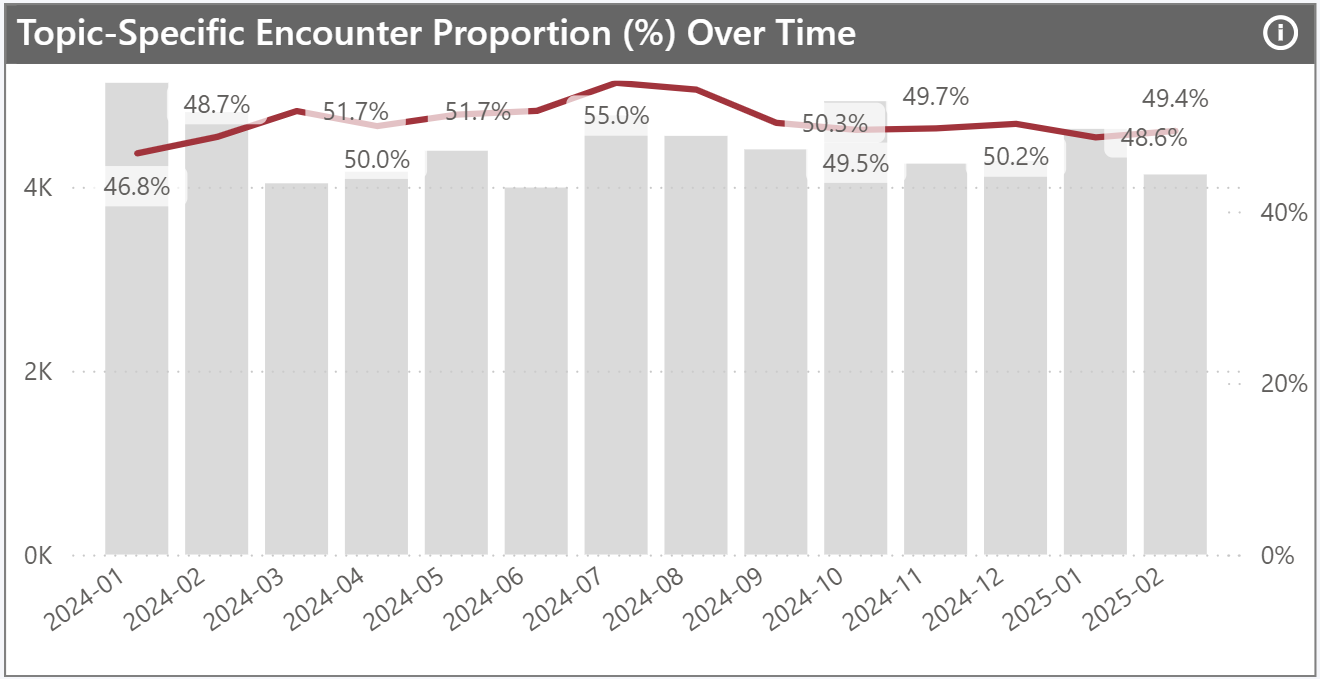}
    \caption*{\textbf{Figure 10.} Non-Clinical Reason Topic Distribution Over Time}
\end{figure}

By tracking staff messages categories over time, the decision support tool enables stakeholders at multiple organizational levels to address critical business questions and operational challenges. Healthcare professionals can analyze how message topic patterns vary across locations and departments to assess equity in access and resource distribution, while identifying rapidly growing subtopics that may indicate emerging patient access challenges suitable for digital self-service solutions to reduce long-term operational costs. PCC managers can identify which subtopics create disproportionate workload burdens and determine where workflow adjustments, automation, or targeted training might improve efficiency, particularly for recurring non-clinical issues in specific offices. Frontline staff can leverage encounter trends to redirect routine inquiries to self-service platforms, reduce manual workload, and proactively prepare for predictable seasonal demand spikes. This transformation from unstructured text data to unified visualization not only quantifies encounter volumes but contextualizes the nature of patient requests, enabling healthcare professionals to continuously adapt staffing, workflows, and digital tools to meet evolving patient needs and demonstrating the substantial value of combining LLM-driven text classification with intuitive dashboard design to convert raw encounter data into actionable insights.

\section{Conclusion and Future Directions}

This paper presents a multi-stage LLM-based framework for analyzing hospital staff messages, demonstrating the feasibility of transforming unstructured communication into structured and actionable insights. Multiple LLM types, including reasoning, general-purpose, and lightweight models, were evaluated, and the outputs from the best-performing LLM, o3, are used to develop the PCC Encounter Message Dashboard under a HIPAA-compliant workflow. The resulting system enables systematic monitoring of encounter trends, supports navigator training, and provides a reproducible pipeline for transforming routine operational data into healthcare analytics. This work demonstrates the application of LLM-based methods to an underutilized, high-volume data source in hospital operations, offering a scalable framework that supports generalizable and reproducible use across diverse healthcare communication settings. Future research may extend this framework beyond topic modeling and classification to also cover HIPAA violation and protocol breach flagging in addition to dismissive language and inappropriate staff messaging detection, enabling a more comprehensive decision support tool that healthcare professionals can act upon in real-time. Furthermore, future work could explore LLM-based agents that autonomously classify staff messages, for instance, by urgency and topic reason, and then coordinate with other agents to clean, validate, and enrich this data before pushing results into the decision support tool. This approach would allow continuous monitoring while reducing human involvement.

\section{Data Availability Statement}
The data that support the findings of this research study contain Protected Health Information (PHI) and cannot be shared publicly due to privacy regulations and ethical restrictions. Inquiries regarding the data can be directed to one of the authors.

\section*{References}
\begin{enumerate}[label={[\arabic*]}, leftmargin=*]
    \item HIMSS. (2020). Digital health: A framework for healthcare transformation. https://www.himss.org/resources/digital-health-framework-healthcare-transformation-white-paper 
    \item Hossain, E., Rana, R., Higgins, N., Soar, J., Datta Barua, P., Pisani, A. R., \& Turner, K. (2023). Natural Language Processing in Electronic Health Records in relation to healthcare decision-making: A systematic review. Computers in Biology and Medicine, 155, 106649. https://doi.org/10.1016/j.compbiomed.2023.106649 PubMed
    \item Mascio, A., Kraljević, Ž., Bean, D., Dobson, R., Stewart, R., Bendayan, R., \& Roberts, A. (2020, May). Comparative analysis of text classification approaches in electronic health records. arXiv Preprint arXiv:2005.06624. arXivResearchGate
    \item Ghassemi, M., Naumann, T., Schulam, P., Beam, A. L., Chen, I. Y., \& Ranganath, R. (2018). A review of challenges and opportunities in machine learning for health. arXiv Preprint arXiv:1806.00388. arXivHealthy ML
    \item Brown III, W., Balyan, R., Karter, A. J., Crossley, S., Semere, W., Duran, N. D., Lyles, C., Liu, J., Moffet, H. H., Daniels, R., McNamara, D. S., \& Schillinger, D. (2021, January). Challenges and solutions to employing natural language processing and machine learning to measure patients’ health literacy and physician writing complexity: The ECLIPPSE study. Journal of Biomedical Informatics, 113, 103658. https://doi.org/10.1016/j.jbi.2020.103658 PubMed
    \item Zhuang, M., Concannon, D., \& Manley, E. (2022). A framework for evaluating dashboards in healthcare. IEEE Transactions on Visualization and Computer Graphics, 28(4), 1715–1731.
    \item He, J., Baxter, S. L., Xu, J., Xu, J., Zhou, X., \& Zhang, K. (2019). The practical implementation of artificial intelligence technologies in medicine. Nature Medicine, 25(1), 30–36.
    \item Heisey-Grove, D., Rathert, C., McClelland, L. E., Jackson, K., \& DeShazo, J. (2021). Classification of patient- and clinician-generated secure messages using a theory-based taxonomy. Health Science Reports, 4(2), e295.
    \item Kalabikhina, I., Moshkin, V., Kolotusha, A., Kashin, M., Klimenko, G., \& Kazbekova, Z. (2024). Advancing semantic classification: A comprehensive examination of machine learning techniques in analyzing Russian-language patient reviews. Mathematics, 12(4), 566.
    \item Sulieman, L., Gilmore, D., French, C., Cronin, R. M., Jackson, G. P., Russell, M., \& Fabbri, D. (2017). Classifying patient portal messages using convolutional neural networks. Journal of Biomedical Informatics, 74, 59–70.
    \item Klimis, H., Nothman, J., Lu, D., Sun, C., Cheung, N. W., Redfern, J., Thiagalingam, A., \& Chow, C. K. (2021). Text message analysis using machine learning to assess predictors of engagement with mobile health chronic disease prevention programs: Content analysis. JMIR mHealth and uHealth, 9(11), e27779.
    \item Ren, Y., Wu, Y., Fan, J. W., Khurana, A., Fu, S., Wu, D., Hongfang, L., \& Huang, M. (2024). Automatic uncovering of patient primary concerns in portal messages using a fusion framework of pretrained language models. Journal of the American Medical Informatics Association, 31(8), 1714–1724.
    \item Chen, X., Ma, W., Li, D., Zhu, F., Routray, S., Guduri, M., \& Margala, M. (2025). Kalman-based adaptive moment estimation optimisation algorithm to enhance GPT in LLMs for medical sentiment analysis of patient health-related feedback. IEEE Journal of Biomedical and Health Informatics.
    \item Sakai, H., Lam, S. S., Mikaeili, M., Bosire, J., \& Jovin, F. (2024). Large language models for patient comments multi-label classification. arXiv preprint arXiv:2410.23528.
    \item Cho, S., Lee, M., Yu, J., Yoon, J., Choi, J. B., Jung, K. H., \& Cho, J. (2024). Leveraging large language models for improved understanding of communications with patients with cancer in a call center setting: Proof-of-concept study. Journal of Medical Internet Research, 26, e63892.
    \item Batugo, A., Hwang, S., Davoudi, A., Luong, T., Lee, N., \& Mowery, D. L. (2024). Textual triage: Assessing GPT-4 for classification of free-text medication-related messages for hypertension management. medRxiv, 2024-09.
    \item Arias, M., Rojas, E., Aguirre, S., Cornejo, F., Munoz-Gama, J., Sepulveda, M., \& Capurro, D. (2023). Defining healthcare KPIs using process mining and patient journey maps. In 2023 XLIX Latin American Computer Conference (CLEI). IEEE. https://doi.org/10.1109/CLEI60451.2023.10345781
    \item Jangland, E., Gunningberg, L., \& Carlsson, M. (2009). Patients' and relatives' complaints about encounters and communication in health care: Evidence for quality improvement. Patient Education and Counseling, 75(2), 199–204. https://doi.org/10.1016/j.pec.2008.10.007
\end{enumerate}

\section*{Appendix}

\begin{figure}[htbp]
    \centering
    \includegraphics[scale=0.22]{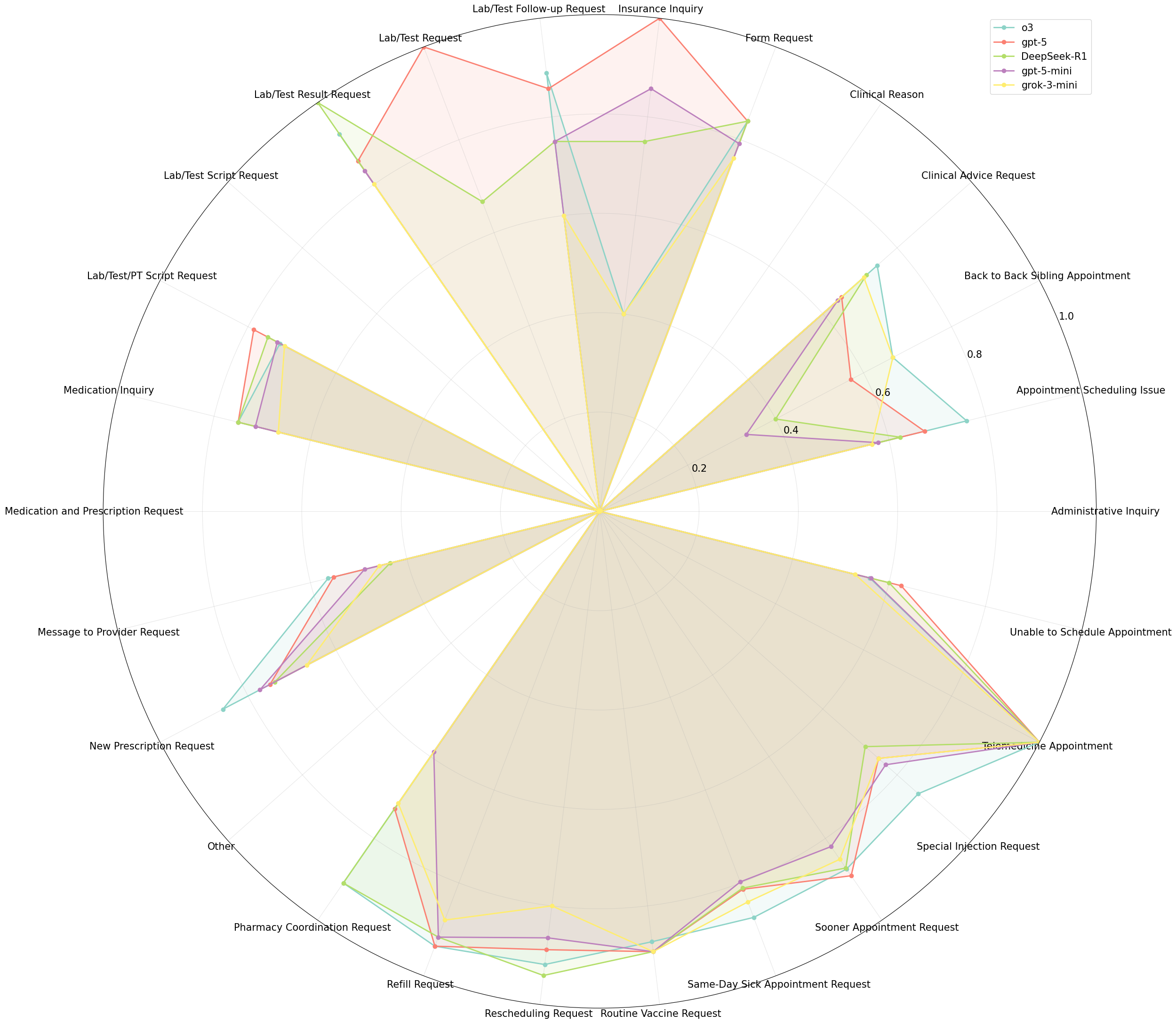}
    \caption*{\textbf{Figure 11.} Top 5 LLMs' Per-class F1-score}
\end{figure}

\begin{figure}[htbp]
    \centering
    \includegraphics[scale=0.27]{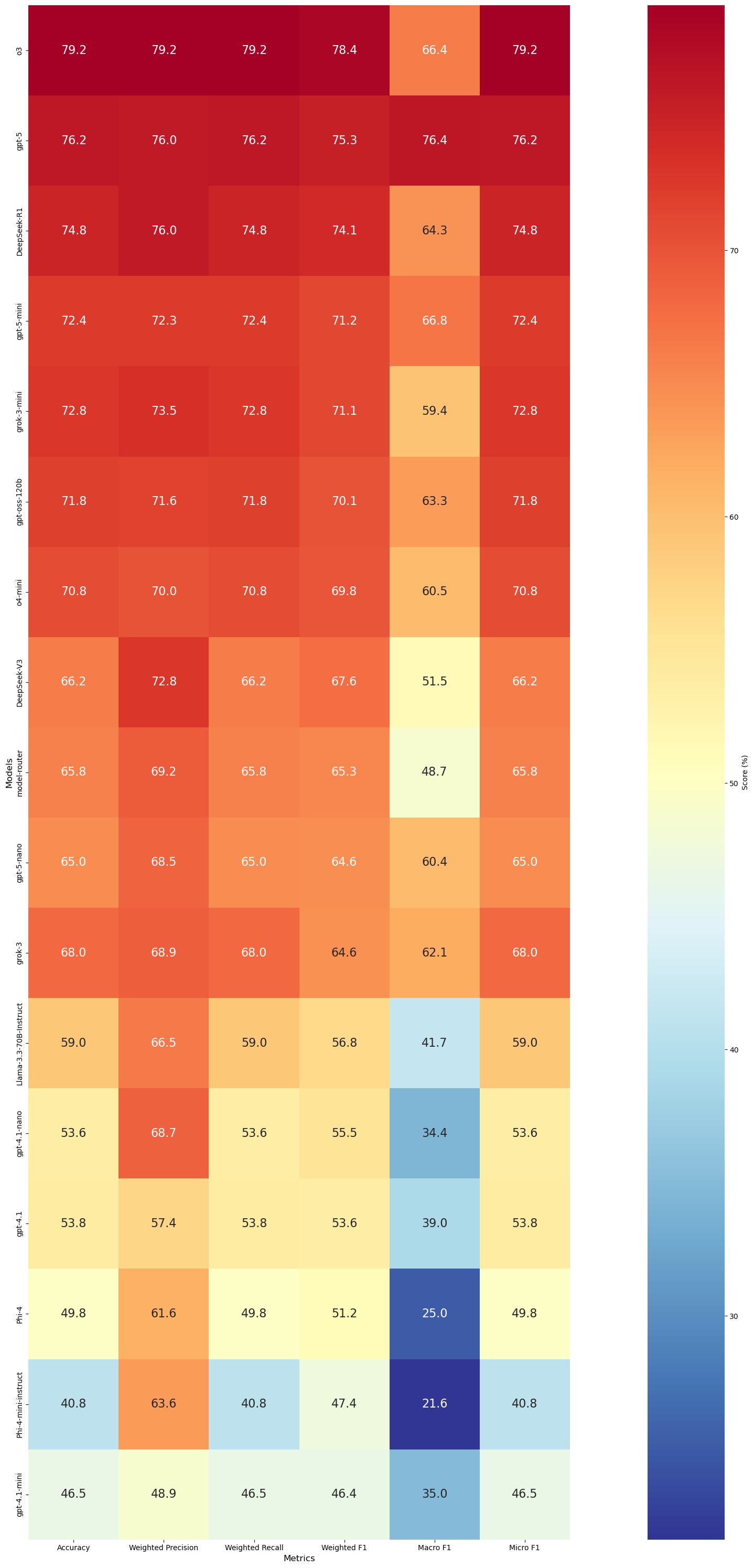}
    \caption*{\textbf{Figure 12.} LLMs' Performance Heatmap}
\end{figure}

\end{document}